\documentclass[a4paper]{article}

\usepackage{INTERSPEECH2020}

\usepackage{amsmath}
\usepackage{graphicx}
\usepackage[colorinlistoftodos]{todonotes}
\usepackage{hyperref}
\usepackage{caption}
\usepackage{subcaption}
\usepackage{booktabs}
\usepackage{longtable}
\usepackage{multirow}

\title{FinChat: Corpus and evaluation setup for Finnish chat conversations on everyday topics}
\name{Katri Leino$^1$, Juho Leinonen$^1$, Mittul Singh$^1$, Sami Virpioja$^2$, Mikko Kurimo$^1$}
\address{
  $^1$Department of Signal Processing and Acoustics, Aalto University, Finland\\
  $^2$Department of Digital Humanities, University of Helsinki, Finland}
\email{katri.k.leino@aalto.fi, juho.leinonen@aalto.fi, mittul.singh@aalto.fi, sami.virpioja@helsinki.fi, mikko.kurimo@aalto.fi}

\begin{document}

\maketitle

\begin{abstract}
Creating open-domain chatbots requires large amounts of conversational data and related benchmark tasks to evaluate them. Standardized evaluation tasks are crucial for creating automatic evaluation metrics for model development; otherwise, comparing the models would require resource-expensive human evaluation. While chatbot challenges have recently managed to provide a plethora of such resources for English, resources in other languages are not yet available. In this work, we provide a starting point for Finnish open-domain chatbot research. We describe our collection efforts to create the Finnish chat conversation corpus FinChat, which is made available publicly. FinChat includes unscripted conversations on seven topics from people of different ages. Using this corpus, we also construct a retrieval-based evaluation task for Finnish chatbot development. We observe that off-the-shelf chatbot models trained on conversational corpora do not perform better than chance at choosing the right answer based on automatic metrics, while humans can do the same task almost perfectly. Similarly, in a human evaluation, responses to questions from the evaluation set generated by the chatbots are predominantly marked as incoherent. Thus, FinChat provides a challenging evaluation set, meant to encourage chatbot development in Finnish.
%Based on the FinChat evaluation set, we also curate a retrieval task to compare the chatbot models.

%%% 19:39
% We use the FinChat evaluation set to compare available state-of-the-art chatbots trained on two out-of-domain Finnish datasets.%

%%%%%%%%% OLD

%Open-domain chatbots capable of casual conversing have the potential to ease communication with computers and improve the quality of life for many.
%Even though the English language now has some resources for open-domain chatbot development, a lack of quality chat data is restricting chatbot development in other languages: models need considerable amounts of domain data for training, and researchers benefit from benchmark data sets to better understand new architectures. Gold standard evaluation sets especially are crucial to creating automatic evaluation metrics for model development.  Otherwise, a comparison of the models would require time and resource-consuming human evaluation. For Finnish, there was no suitable corpus prior to this work. 

%Finnish chat conversation corpus, FinChat, includes real chats on seven topics from people of different ages. In this work, we present an evaluation data set and automatic metrics for Finnish open-domain chatbot evaluation and use it to compare chatbots trained on two out-of-domain data sets. We also create a curated N-choose-k task for retrieval based ranking of models.
%This paper and corpus provide a starting point for Finnish open-domain chatbot research.
\end{abstract}

% TODO:

\section{Introduction}
Recently, open-domain conversational agents or chatbots, capable of casual conversation, have received much attention from NLP researchers. This trend has been supported by regularly organized chatbot challenges like Dialogue State Tracking Challenges\footnote{\url{https://www.microsoft.com/en-us/research/event/dialog-state-tracking-challenge}},  NeurIPS' Conversational Intelligence Challenge\footnote{\url{http://convai.io/}} and Amazon Alexa prize\footnote{\url{https://developer.amazon.com/alexaprize}}. Nevertheless,
training good open-domain chatbots is challenging. The chatbot training requires large amounts of conversational data and an evaluation setup to develop them.
It is often easier to extract conversational data from \textit{online} sources for training than constructing standardized evaluations. Yet, the latter is essential for model development because the alternative is to employ expensive human evaluation for comparing different conversational agents.

Chatbot challenges have overcome this issue by providing standardized evaluation setups to compare models \cite{miller2017parlai, Wolf2019HuggingFacesTS, sedoc2019chateval}. The growth of resources, however, has been restricted to English. For Finnish, like many other languages, there are no chatbot evaluation setups available. Meanwhile, using machine translated corpora from well-resourced languages is dependent on the translation quality which for Finnish is a concern at the moment. In this work, our focus is to bridge this gap for Finnish and to bootstrap open-domain chatbot research in the language. We provide the FinChat corpus and evaluation setup to support this aim.   

% Note: Add MT here. We tried MT but results were poor.
The FinChat corpus consists of Finnish chat conversations on everyday topics collected from voluntary participants. Our goal is to collect conversations that are natural and engaging, which are two important qualities of a human conversation \cite{see-etal-2019-makes}. To ensure naturalness, we do not restrict our participants to a script, specify the language style (formal or informal) or restrict them to specific discussion points. To ensure engaging conversations,
we provide participants with seven broad and diverse topics to guide their conversation.
%we restrict participants with minimal rules and provide them with seven diverse topics. 
Later, the participants self-evaluate each conversation to be engaging or not. 

The FinChat evaluation setup includes a retrieval task to help automatically compare chatbot models. The task provides chatbot with a sentence from a conversation and asks to predict the answer as the continuation from the given list. The task is easy for humans, who achieve $95.1\%$ accuracy, whereas off-the-shelf chatbot models trained on large Finnish conversational datasets perform much worse, barely achieving the accuracy of a random choice, 10\%. We also perform a human evaluation where responses generated by the chatbots to the questions from the FinChat evaluation set are marked for grammatical correctness, intelligibility and coherence. The best generated responses score high, close to original human responses, on grammar and intelligibility but much worse on coherence. Thus, FinChat poses a challenging task for chatbot modelling. To support further research, %These observations suggest FinChat evaluation set to be a challenging one. 
% Bad at Coherence and hits task.
we publicly release the FinChat corpus, our evaluation setup, and training recipes to recreate the considered chatbots at \url{https://github.com/aalto-speech/FinChat}. %We hope these resources will help further research on chatbots in Finnish and the presented collection plan will help support similar efforts in other languages.

\section{Related Work}
Conversational chat corpora in English are mostly knowledge-grounded \cite{topical_dateset, movie_dataset, wiki_dataset} or partly scripted \cite{convai2_dataset}. Conversations are knowledge-grounded by providing participants with a highly specific topic and background reading beforehand. This data generation style results in topical and more coherent conversations. %The generated data helps in studying the impact of longer history and background knowledge in chatbots.
Such data is useful to create chatbots suitable for information retrieval.
For more casual conversations, there are only a few corpora such as PersonaChat \cite{convai2_dataset} that have scripted small-talk conversations. Alternatively, real conversations from social media corpora can be extracted. However, they require significant filtering effort %to remove inappropriate conversations 
to ensure the quality and appropriateness of the content. 
Our approach for conversational chat corpus is to collect diverse casual conversations by not restricting the content and only providing a broad topic as guidance. We also promote diversity by having participants of different ages and giving them the freedom to converse with the conversational language they use in real life. Additionally, we aim for longer conversation by providing participants more time. In most of the above data sets, the length of the conversation are often short with only 5-8 turns except for Topical-Chat \cite{topical_dateset} that has 20 turn conversations. In comparison, the FinChat corpus has conversation length of 14 turns on the average.

Crowd-sourcing is a popular option to gather chat conversations because it gives easy access to a large amount of participants, and the content of the conversation can be controlled for quality. Unfortunately, all languages are not well represented in the crowd-sourcing platforms, and therefore, gathering substantial amounts of data for them is challenging. As this concerns Finnish as well, we created a collection setup and recruited volunteers to have a casual chat with each other. 

% ROUGE METEOR BLEU one of these unnecessary citation?
For chatbot evaluation, using %the most commonly used automatic evaluation metrics are adopted from machine translation and language modelling. P
perplexity, cross-entropy, and translation metrics such as BLEU \cite{papineni2002bleu}, METEOR \cite{banerjee-lavie-2005-meteor}, ROUGE \cite{lin-2004-rouge}, are straightforward to calculate, but show little correlation with human evaluation  \cite{liu-etal-2016-evaluate, lowe-etal-2017-towards}.
%Nevertheless, one recent improvement in the field is the character \textit{n}-gram F-score (chrF) \cite{popovic-2015-chrf}, which shows correlation with human evaluations, and works with morphologically rich languages like Finnish. 
Besides, PersonaChat corpus introduced hits@1/N, which is the success rate of predicting the correct reply out of $N - 1$ random sentences. N-choose-k \cite{shao-etal-2017-generating} could be seen as an extension of hits metric, where it is enough for the correct response to appear in top-k. In our work, we employ these different metrics to evaluate chatbots on the evaluation task.%Depending on the data and values of N and k, an N-choose-k task is surprisingly difficult for humans e.g. in \cite{shao-etal-2017-generating} 10-choose-1 had human performance was 0.45.

In conversation modeling, most recent advances employ Transformer models \cite{NIPS2017_7181, zhang2019dialogpt}. However, many present approaches still use RNN-based encoder-decoder architecture \cite{Su2019, baheti-etal-2018-generating}. In our work, we test off-the-shelf systems from both these two approaches on the FinChat evaluation task. % Aside from architecture, many papers explore the actual method of choosing the best tokens. In \cite{li-etal-2016-diversity} Maximum Mutual Information (MMI) is used as an objective function when generating responses. MMI-based responses are more varied and interesting replies. Another trend is studying the process of decoding the reply. \cite{kulikov-etal-2019-importance} showed sentences generated with greedy search being rated low by humans.% Finally \cite{holtzman2019curious} found both beam search and greedy search generated different word distributions when compared to humans. 

\section{FinChat dataset}
% FinChat, and release the setup to alleviate data collection in other languages
% Challenge with Finnish: crowdsourcing platforms cannot be used (not enough native speakers). Thus made alternative.

%FinChat corpus contains natural, diverse and engaging Finnish chat conversations on seven topics generated by volunteer participants. The corpus was collected to be used as an evaluation set for Finnish conversational models. 
%Conversation are non-scripted. Participants were given broad topics to converse in order not to restrict a natural flow of the conversation. This way, they could steer the content based on their own experience and interests.
%As the style of the language and content of the conversation depends on the background of the speaker, to increase the diversity of the conversations we had participants with different ages and education levels. 
%To ensure that conversations were engaging, each conversation was self-evaluated by the participants.

FinChat corpus provides a set of natural, diverse and engaging conversations on seven topics collected from voluntary participants. To promote natural conversations, we did not provide any scripts to the participants except for the one-word topics. This way, they could steer the content of the conversation based on their knowledge of that topic. In our study, the participants were of different ages and belonged to different academic backgrounds. They also have minimal restrictions on language and conversation style to allow collection of diverse conversation styles. Each conversation is self-evaluated by the participants for engagingness. In this section, we first describe the details on the setup of the collection effort and instructions given to the participants, and then provide essential statistics of the dataset. 

%After each conversation, participants were asked to self-evaluate the conversation they had with a questionnaire which was used to judge the engagingness of the conversation.

%Broad topics for the conversations were chosen to not restrict the natural flow of the conversation. By giving  a general topic, the participants could steer the content based on their own experience and interests which leads to more natural, small-talk-like conversation.

%One of the reasons why chatbots are typically not engaging is due to the conversations in the training data not being engaging either.
%Therefore, we wanted to know which conversations the participants were enjoying.

%The aim was to collect genuine and interesting conversations which user would enjoy having. 

%%%%%%%%%%%%%%%%%%%%%%%%%%%%%%%%%%
%%%%%%%%%%%%%%%%%%%%%%%%%%%%%%%%%%
\subsection{Collection Setup}
For Finnish, crowd-sourcing platforms could not be used due to the lack of native speakers on them. We also tried machine translating PersonaChat, but the results were of poor quality and even incoherent at times. Instead, we setup a chat server\footnote{\url{https://github.com/sdelements/lets-chat}} and invited voluntary participants for the collection effort. The participants were Finnish natives in three age-based user groups: high school students (16-19 years), university students (20-25 years) and university staff (25 years or above).

The data was collected in multiple sessions, where each session had participants from the same group. In each session, participants were paired randomly and used fake names to maintain anonymity.
%The topics also varied in each session.
%Participants used mostly computers but a few chatted with their own smartphones.
%Participants chatted in pairs on assigned channels.
%Pairs were randomly assigned and participants did not know who they were chatting with. 
%Beginning
At the beginning of the session, participants were given a topic to discuss.
% Instructions
They were also instructed to (1) not reveal any personal information, (2) ask one question at the time and wait for their partner's reply, (3) use conversational language, and (4) not use any abusive language.
% Session
After chatting 10-15 minutes, conversation partners were switched and a new topic was given. In a session, each participant had two or three conversations.
% After 
After each conversation, participants self-evaluated their conversation with a questionnaire. The specific questions are reported in Figure \ref{fig:qs_pies}. 
% Cleaning up afterwards
In the case of violating the instructions related to personal information, the data was anonymized following GDPR.

% About topics
%Broad topics were chosen to not restrict the natural flow of the conversation. By giving  a general topic, the participants could steer the content based on their own experience and interests which leads to more natural, small-talk-like conversation.
%Participants were asked to list the topics they talked about in case they did not stick to the given topic.

% Questionnaire

%After each conversation, participants filled a questionnaire about the quality of their conversation. 
%The specific questions and the results of the questionnaire are reported in the following section. 

%%%%%%%%%%%%%%%%%%%%%%%%%%%%%%%%%%
%%%%%%%%%%%%%%%%%%%%%%%%%%%%%%%%%%
\subsection{Statistics}
FinChat corpus contains conversations with message timestamps, sender's id, and metadata information. The metadata includes information on participant id, age group, topic, and questionnaire results. Table \ref{tab:corpus_statistics} shows the conversation statistics for each topic and for each user group. 
The number of conversations on each topic varies because of different number of topics and participants per session. 
The corpus has $86$ conversations with 3,630 messages, 22,210 words with the average word length of $5.6$, and on the average 14 turns per each conversation. % which is typical in conversational Finnish \cite{}. 
% NOTE: Not sure if it is mentioned in any papers but e.g. for forum corpus sample has 5.5 avg word length and news corpus sample around 7 (self computed from sample).

%%%%%%%%% %%%%%%%%% %%%%%%%%% %%%%%%%%% 
%%%%%%%%% TABLE: CORPUS STATISTICS
\begin{table}[t!]
\centering
\caption{FinChat data statistics: the number of conversations (Conv), messages (Mes), and words and the rate of interesting conversations for each topic and group. The groups are university staff, university students and high school students (HS)}
\label{tab:corpus_statistics}
\begin{tabular}{@{}lcccc@{}}
\toprule
Topics & Conv & Mes & Words & Interesting \\ \midrule
Sports & 24 & 1,054 & 5,703 & 77 \% \\
Literature & 15 & 655 & 3,179 & 61 \% \\
TV & 15 & 900 & 4,132 & 71 \% \\
Traveling & 12 & 463 & 4,418 & 83 \% \\
Food & 9 & 240 & 2,140 & 78 \% \\
Movies & 7 & 209 & 1,546 & 57 \% \\
Music & 4 & 149 & 1,263 & 100 \% \\ \midrule
%Groups & & & &
%\midrule
Univ. staff & 41 & 1,526 & 12,700 & 77 \% \\
Univ. students & 10 & 239 & 2,769 & 85 \% \\
HS students & 34 & 1,863 & 6,733 & 66 \% \\ \midrule
All & 86 & 3,630 & 22,210 & 74 \% \\
\bottomrule
\end{tabular}
\vspace{-0.3cm}
\end{table}
%%%%%%%%%%%%%%%%%%%%%%%%%%%%%%%%%%%% 

%%%%%%%%% %%%%%%%%% %%%%%%%%% %%%%%%%%% 
%%%%% GROUP STATISTICS
\begin{table}[t!]
\caption{Group statistics: The number of conversations (Conv), the average word length, the average number of messages in each conversation (Mes / Conv), the number of words in each message (Words / Mes), and the rate of interesting conversations in each age group.}
\label{tab:group_stats}
\centering
\begin{tabular}{@{}lccc@{}}
\toprule
\multicolumn{1}{c}{Groups} & \multicolumn{1}{c}{Word length} & \multicolumn{1}{c}{Mes / Conv} & \multicolumn{1}{c}{Words / Mes} \\ \midrule
Univ. staff & 6.0 & 37.2 & 8.3  \\
Univ. students & 5.5 & 23.9 & 11.6 \\
HS students & 5.0 & 54.8 & 3.6 \\ \bottomrule
\end{tabular}
\vspace{-0.6cm}
\end{table}
%%%%%%%%% %%%%%%%%% %%%%%%%%% %%%%%%%%%

User group statistics are reported in the Table \ref{tab:group_stats}. The majority of the participants were university staff and high school students. It is possible to see interesting differences between these two groups. 
High schools student sent more messages than other groups. However, their messages were a lot shorter and they used smaller words than other participants.
They were also more unsatisfied with the conversations: only $66\%$ reported that their conversation was interesting. In contrast, 
university staff and students rated $77\%$ and $85\%$ of their conversations interesting.
Based on the informal feedback, high school students struggled more on keeping the conversation going, which could partly explain the results. Adults, on the other hand, seemed to enjoy their time talking with others.
%Other reason for could be that the topics might not have been interesting for the students as t
High school students also tended to go off-topic more often than other groups. In the whole corpus, $21.5\%$ of the conversations contained other than the given topic.
%The most popular extraneous topics for them was school or future career.

%Based on our observations the reception from the adults were good and many commented that they had a good time chatting with each other. Students on the other hand found it more difficult to keep the conversation going. However, these were just our observations, we did not ask feedback from everyone. 

\begin{figure}[t!]
\centering
\includegraphics[width=1\columnwidth]{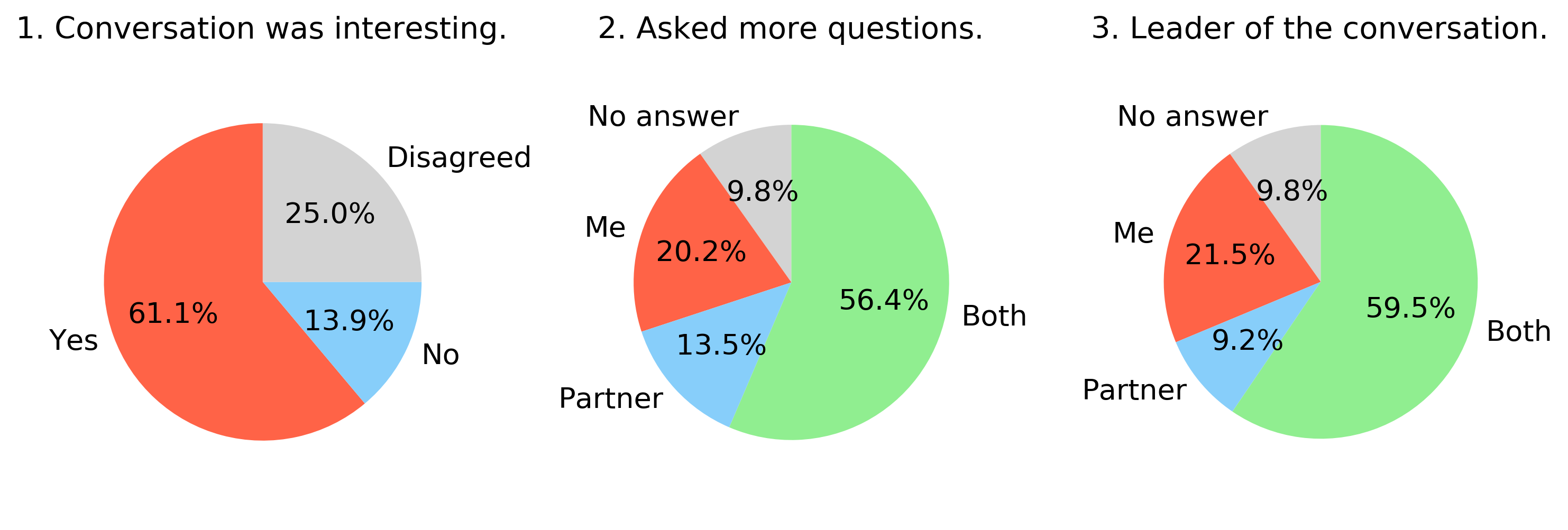}
\vspace{-0.8cm}
\caption{Questionnaire results.}
\label{fig:qs_pies}
\vspace{-0.6cm}
\end{figure}

As we wanted to collect engaging conversations, the participants were asked to evaluate their conversations by filling a questionnaire.
The questions and results of the questionnaire are summarized in Figure \ref{fig:qs_pies}.
$86.5\%$ of the conversations were rated as enjoyable by either one or both the participants, and thus, we were able to collect engaging conversations. 
%Similarly, almost all felt that they were listened to during the conversation. 
%Prior work states the the balance between questions and answers is important for the engaging conversation \cite{see-etal-2019-makes}. Therefore, 
We also asked the participants to answer who asked more questions and who led the conversation. 
%The results for both questions are similar as can be seen from the Figure \ref{fig:qs_pies} the questions $2$ and $3$. 
$68\%$ of the participants answered in the same way for both questions. 
In $84\%$ of the cases, participants agreed who was asking more questions.
However, conversation attendees did not often agree on who led the conversation, as only in $44\%$ of the cases they agreed on this aspect.

%%%%%%%%% %%%%%%%%% %%%%%%%%% %%%%%%%%%
%%%% QUESTIONNAIRE
%%%%%%%%% %%%%%%%%% %%%%%%%%% %%%%%%%%%

\begin{table}[h!]
\centering
\footnotesize
\caption{Evaluation data statistics: the number of conversations (Conv), messages (Mes), and words for each topic.}
\label{tab:eval_statistics}
\begin{tabular}{@{}lccc@{}}
\toprule
Topics & Conv & Mes & Words \\ \midrule
Sports & 4 & 200 & 966 \\
Literature & 2 & 74 & 300  \\
TV & 2 & 77 & 573 \\
Traveling & 2 & 55 & 620 \\
Food & 6 & 140 & 1,311\\
Music & 3 & 103 & 935  \\
\bottomrule
\end{tabular}
\vspace{-0.6cm}
\end{table}
\subsection{Evaluation setup}
Before extracting the evaluation set, we fuse a user's consecutive messages. Then, we select adjacent sentences, which now belong to two different users, that both have more than ten characters. From this set, we select a hundred sentence pairs for every user group. 
% Next sentence needs clarifications: filtered that sentences where it is not clear which sentences come after another. obvious pairs, easily choosable by human evaluators.
Human evaluators inspected these sentence pairs discarding pairs which are not discernible among false alternatives.
The filtered sentence pairs, a total of 226, are utilized for evaluation, as described in Section \ref{ssec:exps}. The corresponding conversations form the evaluation set, details of which are presented in Table \ref{tab:eval_statistics}.

\section{Chatbot evaluation}
In this section, we present the chatbot models used in our study and automatic evaluation metrics to evaluate them. 
We also describe the evaluation setup to compare these models and conduct a human evaluation to understand their limitations.% the limits of are better than others with automatic evaluation metrics, and that the results are in line with human evaluation.

%%%%%%%%%%%%%%%%%%%%%%%%%%%%%%%%
\subsection{Models}

We utilize two popular architectures to train our chatbot models: the encoder-decoder (ED) based model \cite{luong-pham-manning:2015:EMNLP} and the Transformer based model \cite{convai2_dataset,NIPS2017_7181,DBLP:journals/corr/abs-1901-08149}. In the encoder-decoder model, both the encoder and decoder are 2-layered bidirectional gated recurrent units (GRU) \cite{cho-etal-2014-learning} with 500 neurons, with a dropout of 0.2. The decoder applies global attention \cite{luong-pham-manning:2015:EMNLP} with dot alignment function and softmax output layer. Masked negative log-likelihood loss is used with Adam \cite{kingma2014adam} optimizer. The learning rate is 0.0001, and the gradient clipping value is 50. Our Transformer model uses a language modeling task similar to Hugging Face's ConvAI2 submission \cite{convai2_dataset,DBLP:journals/corr/abs-1901-08149}. The encoder part of the model has four layers with 400 neurons, four attention heads and a 0.2 dropout. To predict the actual word, a linear layer with log softmax is applied. The loss function is the negative log-likelihood and the optimizer is Adam with a learning rate of 0.00001 and gradient clipping of 0.5. For chatbot training, we modify the recipes from \cite{lm-example, chatbot-tutorial} to work with subword units generated using Morfessor \cite{DBLP:conf/fsmnlp/CreutzLV05,DBLP:conf/eacl/SmitVGK14}, which are essential for modeling an agglutinative language like Finnish.

We train each of these models on two different Finnish corpora that can be considered out-of-domain with respect to FinChat, but still include conversational language: Suomi24 (S24) corpus \cite{suomi24-2001-2017-vrt-v1-1_fi}, based on extracted messages from a popular Finnish conversation forum, and OpenSubtitles\footnote{\url{http://www.opensubtitles.org/}} (OS) corpus \cite{lison-tiedemann-2016-opensubtitles2016}.  We form a one million sentence subset from the original corpus. % Adam and GRU % unnecessary citations
%Either from Suomi24 corpus \cite{suomi24-2001-2017-vrt-v1-1_fi}, or OpenSubtitles\footnote{\url{http://www.opensubtitles.org/}} corpus \cite{lison-tiedemann-2016-opensubtitles2016}.
%%%%%%%%% TABLE: HUMAN EVALUATION
\begin{table}[!t]
\caption{Mean scores with a standard deviation of human evaluation of question-answer pairs generated by Encoder-Decoder (ED) and Transformer models trained with OpenSubtitles (OS) and Suomi24 (S24) data sets.}
\label{tab:mode-human-eval}
\centering
\begin{tabular}{@{}llll@{}}
\toprule
 & \multicolumn{3}{c}{Human evaluation score} \\
\cmidrule(lr){2-4}
Model & Intelligible & Coherence & Grammar \\ \midrule
ED OS           & 4.51 $\pm$ 1.19   & 1.83 $\pm$ 0.39   & 4.35 $\pm$ 1.10 \\
ED S24          & 4.10 $\pm$ 0.87   & 1.67 $\pm$ 0.19   & 3.95 $\pm$ 0.97 \\
Transformer OS  & 2.38 $\pm$ 0.92   & 1.28 $\pm$ 0.43   & 2.57 $\pm$ 0.66 \\
Transformer S24 & 1.95 $\pm$ 0.07   & 1.33 $\pm$ 0.20   & 2.03 $\pm$ 0.34 \\
Human           & 4.97 $\pm$ 0.70   & 4.85 $\pm$ 0.75   & 4.47 $\pm$ 0.86 \\ \bottomrule
\end{tabular}
\vspace{-0.6cm}
\end{table}
%%%%%%%%%%%%%%%%%%%%%%%%%%%%%%%%
%%%%%%%%% TABLE: MODEL RESULTS
\begin{table*}[!t]
\caption{Results of automatic metrics for Encoder-Decoder (ED) and Transformer models trained with OpenSubtitles (OS) and Suomi24 (S24) data sets. For each training set, the models are evaluated on the corresponding development set and the FinChat evaluation set. }
\label{tab:mode-auto-eval}
\centering
\begin{tabular}{llllllllll}
 &  & \multicolumn{4}{c}{Development set (S24 or OS)} & \multicolumn{4}{c}{FinChat evaluation set} \\
\cmidrule(lr){3-6}\cmidrule(lr){7-10}
Model & Train data & CE & chrF & hits@1/10 & 10C5 & CE & chrF & hits@1/10 & 10C5 \\ \toprule
\multirow{2}{*}{Transformer} & S24 & 0.847 & 0.132 & 0.126 & 0.532 & 1.143 & 0.104 & 0.0619 & 0.469 \\
 & OS & 0.701 & 0.132 & 0.100 & 0.499 & 1.36 & 0.0889 & 0.0664 & 0.527 \\ \midrule
\multirow{2}{*}{\begin{tabular}[c]{@{}l@{}}Encoder-\\ Decoder \end{tabular}} & S24 & 1.07 & 0.0943 & 0.141 & 0.607 & 1.30 & 0.0787 & 0.0973 & 0.540 \\
 & OS & 0.993 & 0.0813 & 0.103 & 0.518 & 1.53 & 0.0554 & 0.0841 & 0.496 \\
\bottomrule
\end{tabular}
\vspace{-0.5cm}
\end{table*}
%%%%%%%%%%%%%%%%%%%%%%%%%%%%%%%%
\subsection{Experiments}
\label{ssec:exps}
We setup a prediction task where the first sentence of an evaluation sentence pair is fed to the model. The output can then be evaluated using character-based cross-entropy averaged over all the next sentences (CE), character n-gram F-score (chrF) \cite{popovic-2015-chrf}, hits@1/N, and N-choose-k \cite{shao-etal-2017-generating} with $N=10$ and $k=5$ (10C5).
We calculate the CE for the next sentence in the pair and average it across all pairs. The chrF score \cite{popovic-2015-chrf} compares the model-generated next sentence with the correct sentence on a character n-gram basis. %This metric has been shown to work better than the more commonly used BLEU score for a morphologically rich languages like Finnish \cite{popovic-2015-chrf}. 
For hits@1/10 and 10-choose-5, we use the pair's first sentence as the question and create a possible answer set by mixing the pair's correct next sentence with randomly chosen nine other sentences from the evaluation. 
% replace clear, correct with the same term as used in the 3.3
24.7\% generated questions did not have a clear, correct answer and were removed manually. 
Given the question, the chatbot chooses from the answer list.  For ranking and predicting the sentences in the list, we use their cross-entropy value assigned by chatbot given the question.
%and the final set was tested by humans resulting $95.1\%$ accuracy. On this 
%to with the second sentence we rank possible next sentences based on their cross-entropy values given the first sentence.  
%The sentence pairs from the FinChat evaluation set was used to form
%, with chosen sentences for 
%hits@1 and N-choose-k tasks. The first sentence if 
%%, was generated from the FinChat corpus. 
%24.7\% generated questions did not have a clear correct answer. These questions were removed manually and 

Humans tested the question and answer set with $95.1\%$ accuracy. According to feedback from human evaluators, some considered the task challenging regardless of the high accuracy, and many had to think of the context and use style cues to deduct the correct sentence. From both Suomi24 and OpenSubtitles, we separated a development set from a held-out set and correspondingly generated one thousand question-answer pairs as FinChat. 
%with a held-out set from each of the corpora used for training. 
%All models were fine-tuned on the development set generated from the corpus used to train them.

Ten human evaluators also evaluated the chatbot models. They were shown ten questions and were asked to score the model-generated answers for each question on three metrics: 1)
\textit{intelligible}: the answer is an understandable sentence in some context, 2) \textit{coherence}: sentence answers the question and 3) \textit{grammar}: the sentence is the grammatically correct form.
The standard scale from 1 (very poor) to 5 (very good) was used. Original answers were rated in the same manner.

% They were asked to rate ten question-answer pairs of four models and original answers by three metrics: (1) \textit{intelligible}: the answer is an understandable sentence in some context, (2) \textit{coherence}: sentence answers the question, (3) \textit{grammar}: sentence is in grammatically correct form. Standard scale from 1 (very poor) to 5 (very good) was used. %The question-answer pairs were selected from the evaluation set so that the question message had only one topic of discussion, for which the answer message clearly replied to.

\subsection{Results}

According to human evaluation scores in Table \ref{tab:mode-human-eval}, encoder-decoder models surpass transformers in every metric, with the model trained on OpenSubtitles data being marginally better than the one trained on Suomi24. The evaluation also suggests that encoder-decoder models can generate intelligible and grammatically correct sentences, but they do not predict coherently based on the previous message. The ED model trained with OpenSubtitles received the best scores among all the models.
On the other hand, transformer models perform poorly in every human evaluation metric: they often produced unintelligible answers and nonsense words.

Despite the problems in text generation, the Transformer models are competitive with encoder-decoder models in terms of the automatic evaluation metrics based on development sets. These results are shown in Table \ref{tab:mode-auto-eval}. %, the transformer models seem competitive with encoder-decoder models, especially with chrF. 
Cross-entropy results suggest that the Transformers had learned the domain of the training data better. They are also better in cross-entropy and chrF for the FinChat data. In contrast, the metric that shows encoder-decoders as better is hits@1/10. With Transformers, the correct reply is less probable than by chance, suggesting they have learned a wrong model of a conversation. While encoder-decoders do not produce coherent reactions to previous messages either,
%do not seem to understand human conversation either,
the probability distribution they have modeled might be more accurate. In 10-choose-5, there might be so many wrong sentences that the correct one easily ends up at the top half, but with hits@1/10, some learning needs to take place.

%As for other automatic metrics, the Transformers might have generated sentences composed of common tokens to score well on them. For the failure at hits@1/10, in a natural chat the response might not be that conditioned on the question as in our domain sets.

\section{Discussion}
%Discussion.

While this paper had success with hits@1/N evaluation metric, the problem of automatically evaluating chatbots is far from solved, and we will continue to develop better automatic metrics for chatbot evaluation. The hits@1/N results showed a clear difference between encoder-decoders and Transformers, which suggests that curated metrics are valuable. Generating the evaluation set automatically, and then using automatic metrics with it, does not seem feasible at the moment.
%Relying on a fully automated process that involves generating the evaluation set itself with the relevant metrics, does not seem feasible at the moment. 

FinChat is a challenging data set because of the free nature of the conversations. 
%A conversation contains multiple successive messages from the same person and 
The messages are not strictly organized as question-answer pairs, as it is common in the more restricted and scripted chats. The messages do not always answer to the previous message but may refer to statements in the conversation history.
%However, the corpus represents realistic and engaging conversations, which benefit the development of new evaluation metrics.

% Future work
Recruiting volunteers to generate chat conversation instead of funded crowd-sourcing is difficult and time-consuming. However, we managed to collect a corpus with size adequate to be used as an evaluation set. Unfortunately, a lot larger data set would be needed for training chatbot models.
In the future, we will continue expanding the data set in order to provide also training material for the models. We will also aim to balance topics and possibly introduce new topics. In addition, we are interested in including older participants to have a more versatile data set, as the way people discuss over chat differs a lot based on their age and background. 
We also aim to include new metrics that would correlate better with human evaluation and measure longer conversation history. New metrics will be necessary, especially when more advanced models are developed.
Furthermore, additional work needs to be put into modeling. Using a much larger pre-trained Transformer model and fine-tune it with the chat corpus is the obvious next step. The current models, both Transformers and encoder-decoders, might also benefit from more thorough hyper-parameter tuning. In addition, more advanced decoding methods have recently shown promising results for increasing coherence and engagingness \cite{kulikov-etal-2019-importance, holtzman2019curious}. Finally, since FinChat has topical information, fusing that to the models is an exciting avenue.
% THESE \cite{kulikov-etal-2019-importance, holtzman2019curious} would go after the word engagingness if there is room

%%%% CONCLUSION IN THE main.tex

%New evaluation metrics to evaluate longer history and conversation have been introduced and while we improve our models, we also aim to include new advanced metrics. 
%

\section{Conclusion}
Other languages aside from English do not have an established evaluation setup for open-domain chatbot research. 
In this paper, we presented Finnish chat conversation corpus, FinChat, and showed that it could be used to evaluate open-domain chatbots. In our experiments, off-the-shelf chatbots based on encoder-decoder and transformer models performed much worse than humans on the FinChat evaluation task. Thus, FinChat posed a challenging problem. We hope these resources will encourage further research on Finnish chatbots and inspire similar efforts in other languages.% and the presented collection plan will help support similar efforts in other languages.
%As an example evaluation compared encoder-decoder and transformer models. Both automatic and human evaluation showed on the FinChat task, off-the-shelf chatbot faced evaluation Encoder-decoder outperformed transformers in both the human evaluation and the automated evaluation metric hits@1/10.
% automated evaluation metric hits@1/10 was able to distinguish the model which was rated better in human evaluation. 
%We hope to see future development for Finnish open-domain chatbot research by releasing the data set, training scripts, and the evaluation setup we presented in this paper.

\section{Acknowledgements}
We would like to thank all volunteers who participated in FinChat data collection and human evaluation studies. This work was supported by the Emil Aaltonen Foundation, Kone Foundation and the FoTran project, funded by the European Research Council~(ERC) under the European Union's Horizon 2020 research and innovation programme~(grant agreement No~771113).
%This work was supported by the %Academy of Finland (grant 329267) and EU's Horizon 2020 research and innovation programme via the project MeMAD (GA 780069) and the Kone Foundation.  
%JL's work has been supported by the Kone Foundation and KL's work by the Emil Aaltonen Foundation.
%SV's work has been supported by the FoTran project, funded by the European Research Council~(ERC) under the European Union's Horizon 2020 research and innovation programme~(grant agreement No~771113).
%The computational resources were provided by Aalto ScienceIT.

\ninept
\bibliographystyle{IEEEtran}
\bibliography{main}

\end{document}